\pdfoutput=1
\documentclass[11pt]{article}

\usepackage{naacl2021}

\usepackage{times}
\usepackage{latexsym}
\usepackage{graphicx}
\usepackage{microtype}
\usepackage{multirow}
\usepackage{booktabs}
\usepackage{bm}
\usepackage[T1]{fontenc}
\usepackage[utf8]{inputenc}

\usepackage{microtype}
\usepackage{enumitem}
\usepackage{todonotes}
\usepackage{xpatch}

\makeatletter
\renewcommand\paragraph{\@startsection{paragraph}{4}{\z@}%
                                    {.25ex \@plus.1ex \@minus.2ex}%
                                    {-1em}%
                                    {\normalfont\normalsize\bfseries}}

\makeatother

\title{Plot-guided Adversarial Example Construction for Evaluating Open-domain Story Generation}

\author{
Sarik Ghazarian,\textsuperscript{\rm 1}
Zixi Liu,\textsuperscript{\rm 1} 
Akash SM,\textsuperscript{\rm 2} \\
\textbf{Ralph Weischedel},\textsuperscript{\rm 1}
\textbf{Aram Galstyan},\textsuperscript{\rm 1}
\textbf{Nanyun Peng}\textsuperscript{\rm 1, 3} \\
\textsuperscript{\rm 1}University of Southern California / Information Sciences Institute \\
\textsuperscript{\rm 2}Indian Institute of Technology Roorkee \\
\textsuperscript{\rm 3}Computer Science Department of University of California, Los Angeles \\
\{sarik, zixiliu, weisched, galstyan\}@isi.edu,
akashsm@ce.iitr.ac.in,
violetpeng@cs.ucla.edu
}

\begin{document}
\maketitle
\begin{abstract}
With the recent advances of open-domain story generation, the lack of reliable automatic evaluation metrics becomes an increasingly imperative issue that hinders the fast development of story generation.
According to conducted researches in this regard, learnable evaluation metrics have promised more accurate assessments by having higher correlations with human judgments.  
A critical bottleneck of obtaining a reliable learnable evaluation metric is the lack of high-quality training data for classifiers to efficiently distinguish plausible and implausible machine-generated stories. 
Previous works relied on \textit{heuristically manipulated} plausible examples to mimic possible system drawbacks such as repetition, contradiction, or irrelevant content in the text level, which can be \textit{unnatural} and \textit{oversimplify} the characteristics of implausible machine-generated stories. 
We propose to tackle these issues by generating a more comprehensive set of implausible stories using {\em plots}, which are structured representations of controllable factors used to generate stories.  Since these plots are compact and structured, it is easier to manipulate them to generate text with targeted undesirable properties, while at the same time maintain the grammatical correctness and naturalness of the generated sentences. 
To improve the quality of generated implausible stories, we further apply the adversarial filtering procedure presented by \citet{zellers2018swag} to select a more nuanced set of implausible texts. Experiments show that the evaluation metrics trained on our generated data result in more reliable automatic assessments that correlate remarkably better with human judgments compared to the baselines.
\end{abstract}

\section{Introduction} \label{section:introduction}

The surge of downstream applications for open-domain natural language generation (NLG), such as dialog systems~\cite{zhang2019dialogpt} and story generators~\cite{rashkin2020plotmachines} necessitates  \textit{automatic} evaluation metrics for quality assessment. 
The existence of accurate automatic evaluation metrics can accelerate the development cycle by facilitating the process of model comparison and hyper-parameter search.
% As the name implies, automatically evaluating open-domain NLG is challenging due to its open-ended nature. 
Many existing reference-based approaches such as BLEU \citep{papineni2002bleu} or ROUGE \citep{lin2004rouge} fail to correlate well with human judgment in open-domain settings due to the fact that there can be potentially many plausible generations that do not have significant overlap with the limited set of given references. This failure invites research on more sophisticated and reliable evaluation metrics.
%for open domain NLG. 

\begin{figure}[t]
\centering
\includegraphics[width=\linewidth]{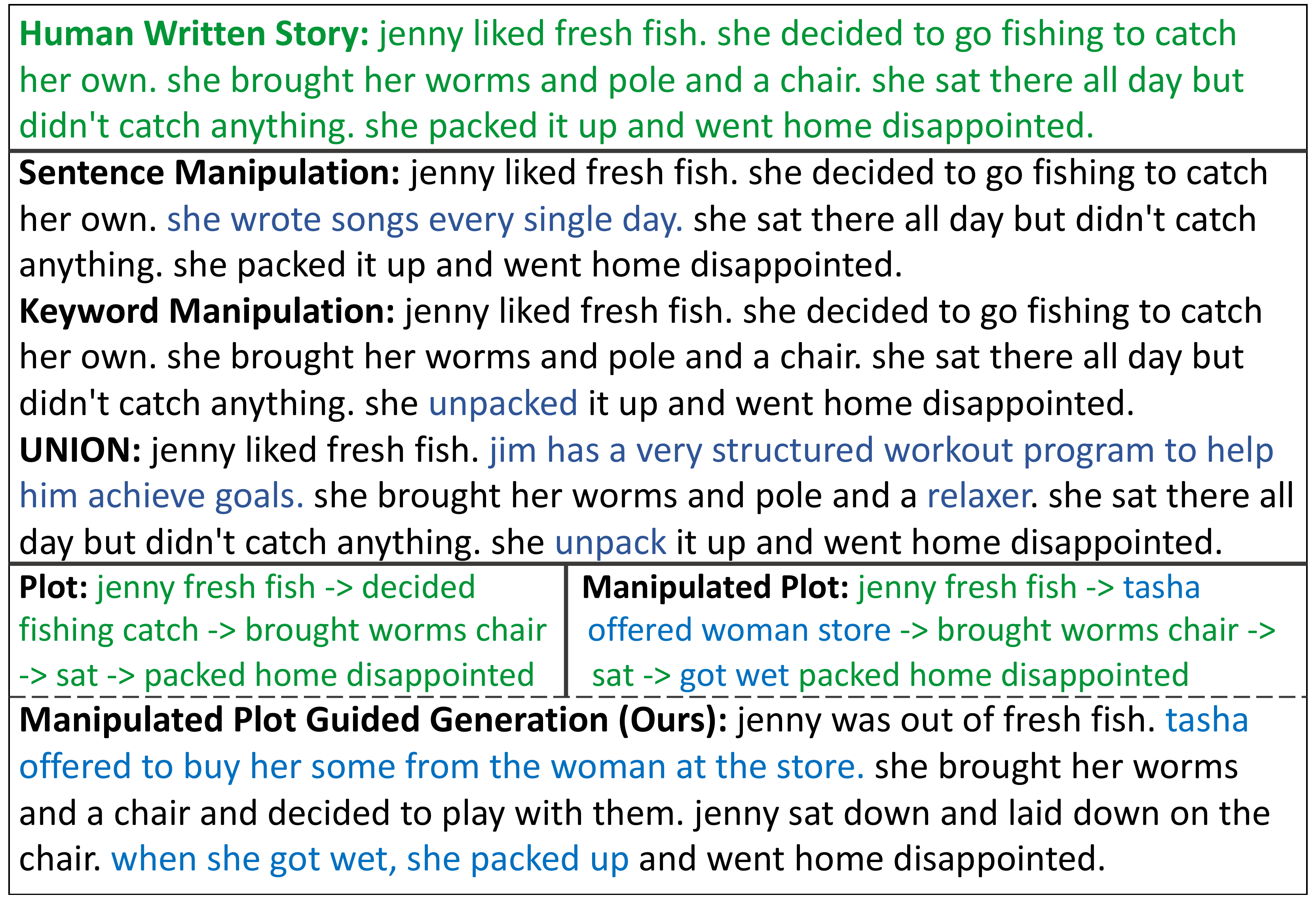}
\caption{Heuristically generated implausible stories (the second block) for a given human-written story (the first block) using sentence, keyword and UNION manipulations versus injecting implausible sources into the story plot (the third block, from the left plot to the right one) and generating a more natural implausible story (the last story). Blue highlights show the implausible sections.}
\label{mot_img}
\vspace{-1.5em}
\end{figure} %\violet{let's change the term from implausible to implausible throughout. Also, can you make the font bigger in your example? Also, for the last example, can you show the original and the manipulated plots? Also, can you show a random sample of the generated text and the adversarially filtered example? You can use a thicker line to separete the heuristic section and the plot-manipulated section.}

%One important aspect of open-domain NLG assessment is \textit{coherence} measurement \citep{newman2010automatic, li2016neural, lai2018discourse}, which indicates to what extent a discourse can convey its meaning to the readers.
%Some existing approaches for automatic evaluation of open-domain NLG reduce scoring to the use of entities \citep{miltsakaki2004evaluation, lapata2005automatic} or topic words \citep{newman2010automatic}, that limits the scope of the evaluation.

Recently, learning-based approaches have been proposed to overcome this limitation by training classifiers to distinguish between plausible and implausible texts \citep{li2016neural, holtzman2018learning}. The choice of training data for learning such classifiers is a key determinant of the metric effectiveness. Existing works take human-written texts as plausible (positive) examples, while the negative samples are heuristically generated by randomly substituting keywords or sentences (See Figure \ref{mot_img}) \citep{li2016neural, Guan2020UNIONAU}. \citet{Guan2020UNIONAU} further improved the quality of evaluators by applying heuristic rules such as adding repetition, reordering and negation (See the UNION story in Figure \ref{mot_img}). 

In this work, we hypothesize that heuristically generated data cannot adequately reflect the characteristics of the implausible texts generated by language models, thus result in suboptimal trained evaluation metrics. 
This deficiency can be mitigated by generating high-quality implausible examples that are closer to the test data. Toward this goal, we propose an approach based on the manipulation of {\em plots}, which are high-level structured representations of generated texts originally used as a content-planning tool for better text generation~\cite{fan2019strategies, goldfarb2020content}. Specifically, we propose to manipulate plots by injecting incoherence sources into them. The generation models conditioned on such manipulated plots lead to implausible texts that have pertinent similarities with implausible machine-generated texts and thus can serve as good negative examples for training evaluation metrics.

%The negative implausible texts construction is not limited to this stage. 

We further improve the quality of training data by incorporating the adversarial filtering technique proposed by \citet{zellers2018swag} to select more challenging negative samples generated from the manipulated plots (See Figure \ref{mot_img}). Eventually, these samples result in more reliable evaluation metrics. The contributions of this work are four-fold:
%\citet{Guan2020UNIONAU} recently proposed a framework to generate low quality texts for assessment purpose. Their approach includes random substitution of keywords and sentences plus inserting repetitive tokens and changing the order of sentences and negating them.

% In this work, we show the inadequacy of heuristically generated negative samples in open-domain NLG evaluation. Indeed, these data oversimplifies and includes a very small portion of feasible implausible generated texts which causes evaluation models fail to correlate well with human judgements. 

% To overcome this issue and generating higher quality implausible texts that are closer to machine generated implausible texts, we examine and show the efficiency of adversarially filtered commonsense inference datasets (SWAG \citep{zellers2018swag}, HellaSwag \cite{zellers2019hellaswag}) on coherence evaluation of open-domain generated texts that encompasses a broader set of implausible examples (such as the second example of Figure (\ref{mot_img})). We leverage this technique to construct a wider range of in-domain implausible texts and show its affirmative role in training models that achieve a significantly higher correlation with human judgments. 

\begin{itemize}[leftmargin=*]
\setlength\itemsep{-.2em}
\item We study the importance of training data for learnable automatic evaluation metrics in open-domain story generation task and show the inadequacy of heuristically generated negative examples in this setting.

\item We propose a novel technique to generate negative samples by introducing plot-level incoherence sources that guide generation models to produce implausible texts. 

\item We show the affirmative role of adversarial filtering techniques in constructing training data for learnable open-domain story generation evaluation metrics. %coherence evaluation task by filtering out a more challenging set of implausible texts generated on manipulated plots.

\item We demonstrate that the evaluation metrics trained on our generated data have a significantly higher correlation with human judgments compared to strong baselines.

\end{itemize}

\section{Related Work} \label{section:related_work}
Existing work on automatic evaluation of generation models can be classified into two subgroups, non-learning-based and learning-based methods, which we briefly summarize below.

\noindent \textbf{Non-learning-based Metrics.}
Some metrics in this group consider the centrality of a text around a specific topic as a proxy for measuring its quality. The transitions of entities in neighbor sentences and their distribution across text have been served as a measurement for quality assessment  \cite{miltsakaki2004evaluation, lapata2005automatic}. Perplexity is another commonly used metric to evaluate the quality of text and story generation models \cite{fan2018hierarchical, peng2018towards}.   

% showed that transitions of entities in neighbor sentences can be a good indicator to score a discourse quality.
% Later, \citet{lapata2005automatic} proposed the entity grid concept to expose how entities are distributed across sentences. They interpreted the joint probability distribution of entities as a plausibility score. \citet{newman2010automatic} measured the quality in terms of topic words similarity. They analyzed diverse scoring methods to obtain the similarity of topic words. All prior metrics reduce the complexity of coherence evaluation to entities and topic words which indeed decrease the accuracy and scope of the evaluation.

\noindent \textbf{Learning-based Metrics.}
%Many researchers resorted to neural networks to take their efficiency in the coherence evaluation task. 
This group of metrics is based on neural-based classifiers trained on a set of positive (plausible) and negative (implausible) texts. The common point between these metrics is using random sentence substitution to construct training examples, while the architectures are slightly different. \citet{li2016neural} trained a neural network with a sigmoid function on top of sentence embeddings extracted from LSTM. 
%to specify the coherence score.  
\citet{lai2018discourse} designed SENTAVG that gets the sentence vectors from LSTM, takes the average of these vectors to represent the whole text, and then passes it through a hidden layer.  

Recently, \citet{Guan2020UNIONAU} proposed a more accurate automatic evaluation metric called UNION. This metric achieved better performance by using BERT \cite{devlin2018bert} as a more effective classification model and have a broader set of negative samples coming from different heuristics. For all learning-based metrics, the simplicity of heuristically generated data samples makes them inadequate for an accurate evaluation of plausibility in open-domain generated texts.

\section{Implausible Text Construction} \label{section:data}

%We formulate the evaluation of open-domain story generation as a binary classification task to distinguish plausible/implausible generated stories, also referred to as positive/negative examples. The existence of high-quality positive/negative examples is obligatory to train accurate and generalizable metrics. Human-written texts have high enough quality to be considered as positive examples. The main challenge arises with the shortage of corresponding negative samples. One possible solution is using unconstrained language models to generate negative samples. However, the does not guarantee correspondence between positive and negative examples. Also, not all generated samples by language models have low quality.  

We formulate the evaluation of open-domain story generation as a binary classification task where the goal is to distinguish plausible and implausible generated stories, also referred to as positive and negative examples. Clearly, the availability of high-quality positive and negative examples is essential for training reliable and generalizable metrics. While human-generated stories can be considered as positive examples, what constitutes good negative examples is a non-trivial question. Specifically, consider a hypothetical decision boundary that separates positive and negative stories. While any point on one side of the boundary will be a negative example, intuitively we want examples that are not too far away from that boundary. To achieve this, we will start from positive examples, and modify them in a controllable manner to generate corresponding negative samples.

%Indeed, any text generated by a low quality generator (or even a random sequence of words) will automatically be a negative example. However, it will not be a high quality negative example, because it will be to classify as such. This intuition suggests that we want negative examples that are on the other side of the hypothetical decision boundary 

%What should we consider as a high quality negative example? The main challenge arises with the shortage of corresponding negative samples. One possible solution is using unconstrained language models to generate negative samples. However, the does not guarantee correspondence between positive and negative examples. Also, not all generated samples by language models have low quality.  

%\violet{I think we should emphasize that the negative examples and the positive examples need to be paird somewhere in this paragraph. Otherwise, people may ask why don't we simple have an unconstrained LM to generate a bunch of negative examples (the reason being: not all the LM generated examples are bad, and we want the negative examples to be correspondent with the positive examples.)} In this section, we explain different approaches to construct such samples.

\subsection{Heuristic Negative Samples} %\violet{you should add a transitional summarization paragraph here.}
There are some widely-used approaches to heuristically manipulate positive examples and change their structure to generate negative examples. 
% Many researchers resort to heuristically generating negative samples for coherence evaluation task \citep{li2016neural, holtzman2018learning, Guan2020UNIONAU}.
\begin{figure*}[t]
\centering
\includegraphics[width=.95\linewidth]{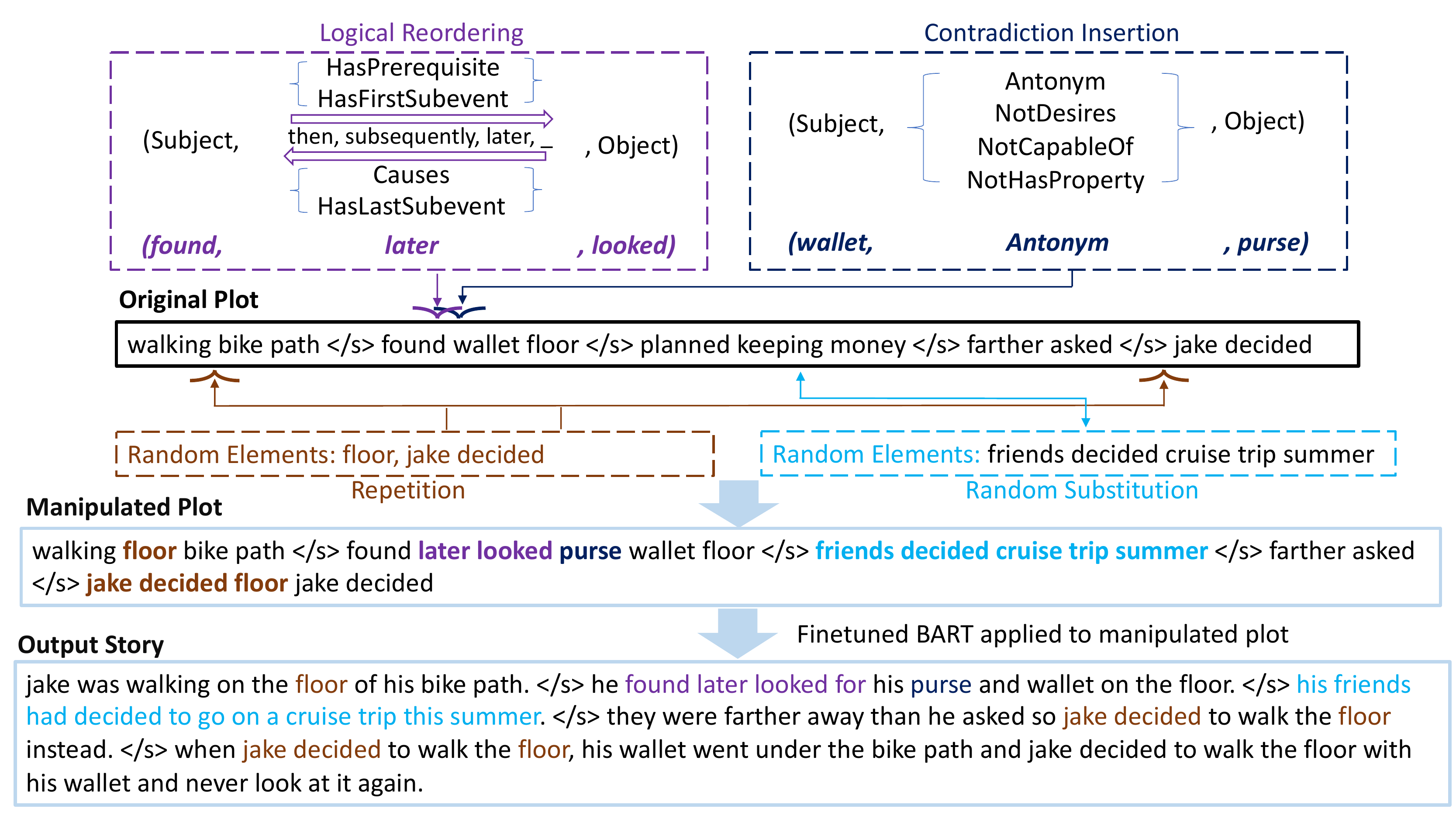}
\vspace{-.5em}
\caption{The plot-level manipulations applied to a human-written story's plot (presented in the black box) to create an implausible story. Each dotted box with a specific color shows a distinct manipulation technique. The manipulated plots are passed through a generation model to generate implausible samples for the evaluation task.}
\label{manipul_plt}
\vspace{-1em}
\end{figure*}
\paragraph{Sentence Substitution.}
Sentence substitution (briefly \textsc{Heur\_Sent\_Sub)} replaces a fraction of sentences in the plausible text with random ones (See Figure \ref{mot_img}). This breaks the discourse-level coherence, making a story not interpretable \citep{li2016neural, holtzman2018learning}.

\paragraph{Keyword Substitution.}
\citet{Guan2020UNIONAU} proposed to apply random substitutions at the keyword-level (briefly \textsc{Heur\_Key\_Sub)}, where a fraction of keywords are randomly substituted with their corresponding antonyms from a commonsense knowledge base such as ConceptNet \citep{speer2012representing} to corrupt the plausibility in the text. ConceptNet consists of (\textit{object}, \textit{relation}, \textit{subject}) triplets. For each selected keyword that exists as an object or subject in the ConceptNet, its counterpart is extracted from one of the contradiction-type relations; \textit{Antonym}, \textit{NotDesires},
\textit{NotCapableOf}, or \textit{NotHasProperty}. For instance, \textit{packed} word in the second example of the implausible text in Figure \ref{mot_img} is substituted by its antonym \textit{unpacked}. %to make it implausible. 
%Keywords without antonyms are replaced with random keywords that have the same part-of-speech tags.
%In the second implausible text of Figure \ref{mot_img}, \textit{stopped} word is substituted by its antonym \textit{continued} to make it less comprehensible.

\paragraph{UNION Manipulations.}
Alongside the keyword and sentence substitutions, \citet{Guan2020UNIONAU} proposed to use repetition, reordering, and negation techniques to generate a more complete and nuanced set of implausible examples. The sentences and keywords are repeated throughout the text to reflect the repetition issue of language models. The order of sentences is changed and negation words are added to make texts implausible due to wrong causal dependencies and conflicted logic. They simultaneously apply some of these techniques to human-written texts to construct negative examples (See third negative story in Figure \ref{mot_img}). We refer to this data as \textsc{UNION\_Data}. 
Despite the demonstrated effectiveness of \textsc{UNION\_Data} in open-domain story evaluation, heuristically constructed negative samples are quite far from machine-generated texts, and thus inadequate to represent a broad set of machine-generated implausible texts.

\subsection{Proposed Approach} \label{section:man_plots}
As we stated above, applying heuristic rules at the {\em utterance} level result in negative examples that are usually unnatural and do not reflect the complex characteristics of machine-generated texts. Instead, we propose to introduce perturbations at a more abstract plot level. Namely, we seek to improve the quality of negative samples using plot-controlled generation with adversarial filtering techniques.
% In order to have a higher quality set of negative samples, we propose the following approaches.

\subsubsection{Plot Manipulations}
Studies have shown that high-quality fluent stories can be generated by planning in advance and leveraging lucrative plots \citep{yao2019plan,fan2019strategies,goldfarb2019plan,goldfarb2020content,rashkin-etal-2020-plotmachines,brahman2020cue}. 
\citet{yao2019plan} leverage a sequence of keywords as the plot representation (also called storyline). %proposed a plan-and-write framework that generates stories from given titles by the intervention of story plots as an intermediate step. They first generate plots as the most important words in each sentence of a story and then generate a story based on a plot. %at the beginning or during story generation process by considering previously generated sentences as the context to generate next words in the plot. 
\citet{fan2019strategies} use semantic role labeling tool to extract plots as abstract presentation of stories over actions and entities. %explored plots as abstract presentation of stories over actions and entities by generating a sequence of predicates and arguments. 
Their experiments affirm that plots have positive effects on generating high-quality stories.

Here we leverage this idea for generating implausible texts, by controllable injection of implausibility sources, or perturbations, into the ground-truth plots. The resulting plot-level manipulations
will force the model to reflect applied implausibility in the generated text and will negatively impact the text's plausibility. In contrast to \citet{Guan2020UNIONAU}, our proposed plot-level manipulations (\textsc{ManPlts}) do not directly change the text at the token level instead, we inject incoherence into language at the concept level. The plot-guided generation guarantees the naturalness of generations since it leverages a well-trained conditional language model. The generated samples are also anticipated to be closer and congruous to the machine-generated texts that will be assessed during the inference time. Concept-level incoherence creates implausible factors that guide models to include that implausible sources. Figure \ref{manipul_plt} demonstrates various proposed plot-level manipulations in dotted boxes.\footnote{Our proposed data, trained models and code is released at \url{https://github.com/PlusLabNLP/Plot-guided-Coherence-Evaluation}} All proposed manipulations are described in the following sections. We refer this data as \textbf{ManPlts}.

\paragraph{Non-logically Ordered Plots.}
Logical conflict is one of the sources for implausibility that results from not-logically ordered concepts in the text. While \citet{Guan2020UNIONAU} covered this type of implausibility by changing the order of sentences, we hypothesize that disrupting the logical order at the concept-level is more efficient. %Indeed, sentence reordering itself is a concept-level reordering that destroys the order of verbs of the sentences. According to human logic concepts, specifically verbs, appear in a determined order. As a concrete example, human expect verb \textit{work} to be followed by \textit{get pay} and the counter-wise would not be acceptable. We apply such disarrangements to reflect logic conflicts in the implausible texts.
To accomplish concept reordering, we first randomly choose verbs from the plot and leverage the COMET \cite{bosselut2019comet} model to predict their subsequent events. Then we dislocate the resulted concept pairs. COMET, which is trained on tuples of the form  ( \textit{subject},  \textit{relation},  \textit{object}), can be used to predict an \textit{object} given a pair of \textit{subject} and \textit{relation}. As an example, given the pair (\textit{work}, \textit{causes}) COMET will predict \textit{get pay} to show that work causes to get paid. We focus on COMET relations \textit{HasPrerequisite}, \textit{HasFirstSubevent}, \textit{Causes} and \textit{HasLastSubevent}  that imply ordering. In the first two relations, object should appear before subject, while in the other two the order is reversed. Therefore, subject \textit{work} comes before \textit{get pay} due to the \textit{causes} relation that holds between them. We flip the correct order of concepts and attach them with or without randomly selected connection words such as \textit{then, later, subsequently} to generate implausible texts (the purple box in Figure \ref{manipul_plt}). 

\paragraph{Contradiction Insertion.}
The plausibility of a text is supported in part by the logical and correct relationship between its words. It can be harmed by accompanying words with their antonyms or other conflicting concepts that add contradiction to the text and make it hard to grasp.
In order to add such kind of implausibility, we propose to insert contradictory sources of randomly selected plots in consecutive positions. For each selected plot, we use ConceptNet \cite{speer2012representing} to extract concepts that hold negation relations such as \textit{Antonym}, \textit{NotDesires},
\textit{NotCapableOf}, and \textit{NotHasProperty} with it and insert them as neighbor plots. In the navy blue box of Figure \ref{manipul_plt}, \textit{purse} has been added before \textit{wallet} as its antonym. This guides the generation model to include consecutive contradictory elements in the generated text that harms the coherence of sections and makes it difficult to interpret.  
%We follow the same approach proposed by \citet{Guan2020UNIONAU} and use ConceptNet \cite{speer2012representing} to return plots antonyms. 
%as a well-known  commonsense knowledge base that includes commonsense triples stating the relations between two different concepts. 
%The head concept relates with tail concept in the form of (head, relation, tail). We randomly select plots in each text and consider them as head or tail nodes in the ConceptNet knowledge base.

%and encourages the model to generate not sensible text like the one in the example. 

 \begin{table*}
\begin{center}
\small
\begin{tabular}{@{\ \ }p{4.5cm}@{\ \ }@{\ \ }p{2cm}@{\ \ }@{\ \ }p{8.5cm}@{\ \ }}
\toprule
 \textbf{Manipulated Plots} & \bf Manipulation Type & \bf Generated Implausible Stories  \\\midrule
 
  1. \textit{\textbf{waiting airport cousin jimmie}} & Random\_Subs. & 1. tobias was waiting at the airport to see his cousin jimmie.\\  
  2. \textit{\textbf{ended}} practiced month & Repetition & 2. although his parents ended up being unwanted, he practiced with his dad every day for a month.\\ 
  3. \textit{\textbf{ended unwanted}} wanted parents proud & Contr\_Insert. & 3. his parents ended up being unwanted, but he really wanted to make his parents proud. \\  
  4. ended losing & - & 4. he ended up losing to his cousin jimmie because he didn't practice enough.  \\ 
  5. parents ended proud & - & 5. but his parents ended up being proud of him anyway. \\\midrule
  
  1. ﬁnished \textit{\textbf{friend turned}} assignment & Repetition &  1. i finished my job , and my friend and i turned in our homework assignment.\\
  2. friend & - & 2. my friend and i went to the mall.\\  
  3. \textit{\textbf{vendors games rides}} & Random\_Subs. & 3. we went to vendors, played many games, and had rides.\\ 
  4. turned & - & 4. when we got home, i turned in the assignment.\\
  5. \textit{\textbf{went to class later}} teacher called \textit{\textbf{home}} ofﬁce & Logic\_Reorder. Contr\_Insert. & 5. when we went to class later teacher called us home from the office. \\\midrule

  1. \textit{\textbf{made gina nervous perform}} & Random\_Subs. & 1. gina’s mom made gina get nervous about having to perform.\\
  2.  \textit{\textbf{pained}} ended \textit{\textbf{scared}} missing & Repetition Logic\_Reorder. &2. she was pained that she ended up scared of missing her bus \\ 
  3. bus \textit{\textbf{scared allie}} & Repetition  & 3. she was on the bus and scared of allie.\\ 
  4. one & - & 4. no one was on the bus and she didn’t know where they were.\\ 
  5. \textit{\textbf{allie}} scared & Repetition & 5. therefore gina and allie were too scared to ride on the bus together.\\\midrule

 1. billy noticed \textit{\textbf{billy}} & Repetition  & 1. billy noticed that his buddy billy was out of gas.\\ 
 2. \textit{\textbf{grandpa loved recall}} & Random\_Subs. & 2. billy's grandpa had just loved to recall a car recall he didn't want to recall.\\ 
 3. \textit{\textbf{billy}} ﬁnished ﬁlling \textit{\textbf{drove}} & Repetition & 3. billy got in his car, finished filling it up, and drove away.\\
 4. \textit{\textbf{unsuddenly}} suddenly & Contr\_Insert. & 4. suddenly, suddenly, suddenly, billy's car was out of gas.\\
 5. \textit{\textbf{billy noticed}} billy driven & Repetition& 5. billy noticed later that billy had driven off with that car recall.\\\midrule
  \bottomrule 
\end{tabular}
\end{center}
\caption{Examples of implausible stories generated based on manipulated plots. Bold italic keywords represent manipulated plots resulted from different proposed manipulations shown in the middle column.}
\label{man_plt_examples}
\vspace{-1em}
\end{table*}
 
\paragraph{Repetition Insertion.}
Repetition is one of the common issues that many generative models suffer from. Recently proposed top-k \cite{fan2018hierarchical} and top-p  \cite{holtzman2019curious} sampling techniques partially mitigated but not completely solved this issue.  \citet{Guan2020UNIONAU} proposed to replicate this problem in negative implausible text construction by repeating N-grams in consecutive positions. These heuristically constructed outputs only mirror local repetition issues, while the state-of-the-art generative models produce more complex and subtle repetitions throughout the whole text.

We propose to repeat random plots of each text in various positions that would force the language model to duplicate them throughout the text and exhibit more realistic machine-generated repetitive examples.
In Figure \ref{manipul_plt}, the repetition of \textit{floor} and \textit{jake decided} compels the model to generate boring and repetitive sentences.

\paragraph{Random Substitution.}
Random sentence substitutions employed by many evaluation models amplify the implausibility sources in the text by inserting completely off-topic sentences that could potentially result in topical inconsistency throughout the text. Such scenarios are less likely for state-of-the-art high-quality generation models that use encoded context to generate tokens.
%  Therefore decision-boundary generated by models trained on this data would be far from real decision boundary. In this work, we show that such kind of completely off-topic sentences are not close to implausible texts generated by generation models since even the low quality texts generated by such models are somehow irrelevant to the context and they do not include completely not relevant sentences. These language models usually encode previously seen context to generate each token in each time-step, therefore the scenario of having very offtopic sentence that even sometimes the subject of sentence do not match with the previous parts is not applicable. 

Once again, we propose to do the replacement at the plot level. Within our approach, even though the inserted random plots are completely irrelevant, the model would attempt to incorporate them into the text as much as possible by using encoded context sentences. This can be seen in the third sentence of Figure \ref{manipul_plt}. Even if this sentence's plots are randomly inserted, the model is able to generate a sentence that does not have significant topical inconsistency, thanks to the contextualized nature of the generative process. 

Table \ref{man_plt_examples} depicts four different machine-generated stories, each containing five sentences that are conditioned on the manipulated plots. Bold italic keywords represent manipulated plots resulted from the proposed approaches shown in the middle column.

\subsubsection{Adversarial Filtering}
Adversarial filtering (AF) technique was originally proposed to generate high-quality negative examples for a grounded commonsense inference task \cite{zellers2018swag}. AF uses a committee of trained models to identify more appropriate negative endings from a pool of candidate samples generated for a given context. For each human-written text, there are \textit{N} machine-generated endings. The goal is to select the most unbiased subset (\textit{A}) of generated endings with similar stylistic features to the human-written ones.

AF starts by randomly specifying the best endings in the assignment set (\textit{A}) from all \textit{N} endings of each context \cite{zellers2018swag}. In each iteration, the data is divided into two parts. The first part is used for training a classifier to distinguish high/low quality endings, and the second part is used for replacing easy endings in \textit{A} with adversarial endings from \textit{N}. Easy endings are the ones that a trained classifier assigns a  much lower score compared to human-written texts, e.g., due to their significantly different writing styles. Adversarial texts have a higher positive probability than easy texts indicating the challenge for a classifier to distinguish them from human-written texts. The replacement of easy texts with adversarial ones maximizes the empirical error of the trainable classifier. The steps outlined above are repeated till the assignment set is filled with high-quality endings for each context.

We use AF on top of the plot-based manipulations for generating implausible texts (briefly call \textsc{AF\_ManPlts}). Our approach for negative texts construction has two main stages: 1) generate a set of \textit{N} implausible texts conditioned on manipulated plots 2) pick out the \textit{A} most challenging high-quality implausible texts without stylistic biases based on applied adversarial filtering technique to increase the quality of negative samples. 
% We briefly call the data generated in this manner . 

% from the and use it as training data to learn coherence evaluators. In this setting, for each text we have a context and a positive human written texts for that context and a set of continuing generated texts. In this work, we consider two different types of negative continuations for each text: 1) various machine generated texts conditioned on the context and we do not have any plots to guide the model generate the incherent texts 2)various machine generated texts conditioned on the manipulated plots that have been explained in previous section. Throughout the paper, we call the first group of data as \textbf{AF} while we call it the second group as \textbf{Manipulated\_Plts\_AF}. 

\section{Learnable Evaluation Models} \label{section:models}
We assess the plausibility of a text by training a classification model on the data that consists of human-written texts (positive examples) and constructed implausible stories (negative examples). Binary classifiers trained on this data can produce the probability of plausible/implausible labels for each text. The predicted probability of the positive class is interpreted as the text's plausibility score.

\subsection{Fine-tuning Language Models}
The effectiveness of large pretrained language models has been proven in NLP downstream tasks \citep{devlin2018bert,liu2019roberta,yang2019xlnet, beltagy2020longformer}. RoBERTa introduced by \citet{liu2019roberta} is one of these models achieving impressive performances on text classification. We employ RoBERTa for our plausibility classification task. We start from pretrained RoBERTa parameters and fine-tune them on the constructed evaluation dataset to predict plausibility scores.

One of the main limitations of RoBERTa is its length requirement of at most 512 tokens. %It is not capable of encoding and processing long texts due to the self-attention mechanism which makes it computationally insufficient. 
Recently, this limitation was addressed by considering a sparser set of attention mechanisms such as locality-sensitive hashing and sliding window attentions,  which reduce the computation complexity from $\mathcal{O}(n^2)$ to $\mathcal{O}(n\log{}n)$ and $\mathcal{O}(n)$ respectively \cite{kitaev2020reformer, beltagy2020longformer}. In this work, we broaden the scope of the text plausibility evaluation to cover not only short but also long texts with more than 512 tokens. To this end, we examine and evaluate the quality of long texts using Longformer \cite{beltagy2020longformer} that has linear complexity in terms of the number of tokens in a text. We fine-tune the pretrained Longformer for long text plausibility evaluation. %on the set of plausible/implausible texts.
%\footnote{We fine-tune both Longformer and RoBERTa using \url{https://github.com/huggingface/transformers}.}

\subsection{Baselines}
We benchmark both fine-tuned classifiers on the manipulated data with the two following baselines.

\paragraph{UNION.} Recently, \citet{Guan2020UNIONAU} proposed an automatic evaluation metric by training a BERT model \cite{devlin2018bert} with an auxiliary reconstruction objective which helps to recover the perturbation from a negative sample. The proposed model is trained on negative implausible texts constructed by adopting repetition, substitution, reordering, and negation sampling techniques. This model and its proposed approach for data construction were compared with previously proposed methods and shown to be more efficient.

\paragraph{SENTAVG.}
We complete our investigation by selecting SENTAVG \cite{lai2018discourse} as another baseline model for the plausibility evaluation task.
%It takes the average representation of the text's sentence embeddings resulted from LSTM and passes it through a hidden and softmax layer to get coherence scores.
SENTAVG leverages LSTM to get sentence representation from their words GloVe embeddings. All the sentences vectors are averaged to form the representation for the whole text and this vector is passed to a hidden layer. A softmax layer at the end computes the probability distribution of texts over positive and negative labels.

\section{Experiments} \label{section:experiments}
We investigate the effectiveness of our proposed approach versus heuristic negative sampling techniques by focusing on the evaluation of open-domain story generation models in two datasets with short and long stories. We show the generalizability of metrics trained on our proposed plot manipulation data. We also separately assess the impact of each manipulation technique on the metric accuracy.

\begin{table}
\centering
\small
\begin{tabular}{lc}
\hline
\textbf{Dataset} & \textbf{Train/Valid/Test}\\
\hline
\textsc{Heur\_Sent\_Sub} & 47.1k/5.9k/5.9k\\
\textsc{Heur\_Key\_Sub} & 47.1k/5.9k/5.9k\\
\textsc{UNION\_Data} & 47.1k/5.9k/5.9k \\ 
% Adv\_Flt & 57k/19k/19k  \\ 
\textsc{ManPlts} &  47.1k/5.9k/5.9k \\
\textsc{AF\_ManPlts} & 94.2k/11.8k/11.8k \\ \hline
\end{tabular}
\caption{Plausibility evaluation datasets for ROC stories using different negative sampling techniques.}
\label{tab:data_ROC}
\vspace{-1.5em}
\end{table}

\subsection{Datasets} 
We conduct our experiments on two 
English stories datasets that are significantly different in terms of length and topic; ROCStories (shortly ROC) and Writing Prompt (briefly WP) datasets including on average 49.4 and 734.5 tokens in each story.

\paragraph{ROCStories.}
ROCStories is a resource of five-sentence commonsense stories collected via crowdsourcing~\citep{mostafazadeh2016corpus} covering a logically linked set of daily events. %It includes 78529/9816/9816 train/valid/test stories. 
We follow the approach proposed by \citet{yao2019plan} to extract story plots (storylines) for the stories and manipulate them to guide conditional language models to generate negative samples. %\footnote{\url{https://bitbucket.org/VioletPeng/language-model/src/master/rocstory_plan_write/}}

\paragraph{Writing Prompt.}
Writing Prompt dataset contains abstract high-level prompts and their corresponding long human-written stories from an
online forum \cite{fan2018hierarchical}. %WP consists of overall 272600/15620/15138 train/valid/test stories. 
To apply the plot manipulation technique for implausible text construction, we follow the procedure proposed by \citet{fan2019strategies} to extract the plots with verb and argument type role labeling tags. 
%We post-process plots to retrieve verb and argument type ones and consider them as ground-truth plots. 

\paragraph{Data Preparation.} We split the stories from both datasets into two subsets for training generation and evaluation models, respectively. We use 70 percent of stories in ROC (ROC\_LM) and WP (WP\_LM) for fine-tuning GPT2 \citep{radford2019language} language model with batch size of 4.\footnote{We fine-tune GPT2 language model using \url{https://github.com/huggingface/transformers}.} 
%We fine-tune GPT2 \citep{radford2019language} model to generate open-domain texts for evaluation purposes.
After 3 epochs of fine-tuning, the perplexity on the validation set of ROC and WP datasets are 8.28 and 25.04, respectively.

The remaining 30 percent of stories from ROC (ROC\_Eval) and WP (WP\_Eval) are used for training and evaluating the evaluation models. All stories in the original dataset represent plausible texts. We apply approaches from Section \ref{section:data} to augment negative samples. Table \ref{tab:data_ROC} and Table~\ref{tab:data_WP} summarize the resulting datasets for ROC and WP. 
%resulted from applied approaches.
In \textsc{Heur\_Sent\_Sub}, we extract all stories with at least 2 sentences and replace 50\% of their sentences with random ones. For \textsc{Heur\_Key\_Sub}, we do random substitution of 15\% of keywords with their corresponding antonyms extracted from ConceptNet and ignore stories without substitutable keywords. 
%for heuristically generating negative samples.
The UNION\_Data is resulted by following 
%repetition, substitution, negation, and reordering 
rules from \citet{Guan2020UNIONAU} and is applied to stories with at least four sentences.
% We followed the same negative sampling techniques proposed by \cite{Guan2020UNIONAU} to automatically construct negative samples and assess their proposed approach for creating negatives samples in the training plausibility evaluation metrics. In all these data, one negative sample is added for each positive human-written text resulting a balanced sets of plausible/implausible stories. 
% For Adv\_Flt dataset, we first fine-tune GPT2 on first section of ROCStories dataset. The perplexity of 11.34 is resulted from GPT2 model fine-tuned for three epochs on test set of ROCStories data. This fine-tuned GPT2 model is used to generate six negative samples for each positive human-written text that are filtered by applying adversarially filtering technique to select the most three challenging texts as negative samples.
\begin{table}
\centering
\small
\begin{tabular}{lc}
\hline
\textbf{Dataset} & \textbf{Train/Valid/Test}\\
\hline
\textsc{Heur\_Sent\_Sub} & 163.1k/9.3k/9.1k\\
\textsc{Heur\_Key\_Sub} & 162.7k/9.3k/9.0k\\
\textsc{UNION\_Data} & 161.8k/9.2k/9.0k \\ 
% Adv\_Flt & k/k/k  \\ 
\textsc{ManPlts} &  84.5k/4.7k/4.7k \\
\textsc{AF\_ManPlts} & 107.2k/35.7k/35.7k \\ \hline
\end{tabular}
\caption{Plausibility evaluation datasets for WP stories using different negative sampling techniques.}
\label{tab:data_WP}
% \vspace{0.05em}
\end{table}

To create \textsc{ManPlts} dataset, we first fine-tune the BART model \cite{lewis2019bart} with a batch size of 8 for three epochs on pairs of ground-truth plots and stories from ROC\_LM and WP\_LM data with the resulting perplexity of 3.44 and 6.79 for the validation sets. Afterward, 15\% of plots are employed and two up to four proposed manipulation techniques in Section \ref{section:man_plots} are randomly selected and applied. We leverage the fine-tuned BART model and use the top-50 sampling technique with a temperature of 0.8. We specify the maximum length of 200 for ROC dataset and 1024 for WP dataset to generate implausible texts on manipulated plots. 

In the \textsc{AF\_ManPlts} dataset, we apply the adversarial filtering technique on top of six generated implausible stories using the fine-tuned BART model conditioned on the manipulated plots. The output contains each human-written story and its three most challenging implausible samples.

 \begin{table}[t]
\centering
\small
\begin{tabular}{cccc}
\hline \textbf{Data} & \textbf{Texts} & \textbf{Annotators} & \textbf{Kappa} \\ \hline
ROC & 300 & 27 & 0.61\\
% ROC & 250 & 46 & 20 & 0.83\\
WP & 300 & 75 & 0.56 \\
\hline
\end{tabular}
\caption{Statistics and inter-annotator agreement of AMT annotations for plausibility metrics evaluation.}
\label{amt_stats}
\vspace{-1.5em}
\end{table}

\subsection{Human Annotations}
The performance of automatic evaluation metrics is assessed based on their correlations with human judgments. To this end, we gather human evaluations and examine the Spearman ($\rho$) and Kendall ($\tau$) correlations with metrics predicted scores \cite{newman2010automatic, lai2018discourse, Guan2020UNIONAU}. Spearman and Kendall are beneficial in estimating monotonic associations for not normally distributed and ranked scores.

We collect human judgments through Amazon Mechanical Turk (AMT) experiments.
We randomly choose 150 human-written stories from ROC\_Eval and WP\_Eval test sets and 
150 machine-generated texts by the fine-tuned GPT2 models. Five distinct participants are asked to rate each story on a scale of 0 to 5 (from {\em not at all plausible}  to {\em completely plausible}). 
% We also asked annotators to select reasons that the text is not completely plausible. Not relevant (\textbf{NR}) and nonsensical (\textbf{NS}) indicate texts with disjoint and nonsensical sequences were among the provided set of reasons.\footnote{Human annotations will be provided alongside the code.} 
We prepare an attention check test to guarantee the accuracy of human annotations and recollect evaluations for users who do not pass the test. 
%or have a less than 0.2 agreement with other annotators. 
The average score of the five annotators is treated as the final human score for each text. We normalize human scores to be in the same range of 0-1 as the model’s output scores are. Table \ref{amt_stats} shows the statistics and agreements in the conducted experiments.

\subsection{Experimental Setup}
We conduct a comprehensive set of experiments to examine and show the importance of training data in the plausibility evaluation task. We train both evaluation and language models on a machine with a GeForce RTX 2080 Ti GPU. 

In our experiments, we have SENTAVG as the baseline model. 
%since it has proved its superiority towards other metrics including entity-based models. 
We compare SENTAVG across more powerful classifiers -- RoBERTa for ROC stories and Longformer for WP stories (FT\_LM). We fine-tune pretrained RoBERTa-base model with the learning rate of 2e-5 and batch size 8 for three epochs and process the ROC stories with a maximum of 128 tokens. To evaluate WP with lengthy stories, we fine-tune pretrained Longformer-base model with the learning rate of 2e-5 and batch size 3 by encoding texts with at most 1024 tokens for three epochs.\footnote{We fine-tune RoBERTa and Longformer models using \url{https://github.com/huggingface/transformers}.} 

%SENTAVG can process texts regardless of their lengths due to its LSTM-based architecture. 
We complete the models' comparisons by incorporating the recently proposed UNION model \cite{Guan2020UNIONAU} to our experiments. We retrain it on the ROC\_Eval and WP\_Eval sets with the same hyper-parameters stated in their paper.

\begin{table}[t]
\begin{center}
\small
\begin{tabular}{@{}l@{\ }l@{\ \ }c@{\ \ }c@{\ \ }c@{\ \ }c@{\ }}
\toprule
 \multirow{2}{*}{\textbf{Dataset}} &  \multirow{2}{*}{\textbf{Model}} & \multicolumn{2}{c}{\bf ROC} & \multicolumn{2}{c}{\bf WP } \\ \\  %\fontsize{11}{11} 
 &  & \bm{$\rho$} &  \bm{$\tau$}  & \bm{$\rho$} &  \bm{$\tau$} \\\midrule
  \multirow{2}{*}{\textsc{Heur\_Sent\_Sub}} 
  & \textsc{SENTAVG} & 0.04 & 0.03  & -0.13 & -0.10 \\
  & \textsc{Ft\_LM}  & 0.10 & 0.07 & 0.12 & 0.10\\
\midrule
  \multirow{2}{*}{\textsc{Heur\_Key\_Sub}} 
  & \textsc{SENTAVG} & -0.04  & -0.03  & -0.26 & -0.18 \\
  & \textsc{Ft\_LM} & 0.31 & 0.22 & 0.08 & 0.06\\
\midrule
  \multirow{3}{*}{\textsc{UNION\_Data}} 
   & \textsc{SENTAVG} & 0.11  & 0.08  &  -0.22 & -0.15\\
   & \textsc{Ft\_LM}  & 0.46 & 0.34 &  0.49  & 0.32\\ 
   & \textsc{UNION} & 0.22  & 0.15  & 0.19 & 0.15 \\
\midrule
  \multirow{2}{*}{\textsc{ManPlts}}  
   & \textsc{SENTAVG} & 0.24 & 0.16  & 0.22 & 0.20\\
   & \textsc{Ft\_LM}   & 0.50  &  0.37  &  0.71 & 0.48\\   
\midrule
  \multirow{2}{*}{\textsc{AF\_ManPlts}}  
   & \textsc{SENTAVG} & 0.22 &  0.16  & 0.25 & 0.23\\
   & \textsc{Ft\_LM} & \bf 0.56 & \bf 0.41 & \bf 0.74 & \bf 0.52\\ 
  \bottomrule 
\end{tabular}
\end{center}
\caption{Higher correlations of plausibility evaluation models trained on manipulated plots and adversarially filtered negative samples with human judgments versus heuristically constructed negative samples. Ft\_LM represents fine-tuned RoBERTa and Longformer models for ROC and WP datasets, respectively. }
\label{models_corr}
\vspace{-1.5em}
\end{table}

\subsection{Experimental results}
Table \ref{models_corr} depicts the quantitative results of correlation analysis between human and automatic evaluation metrics. For almost all constructed datasets for evaluation, the RoBERTa and Longformer in the case of short and long stories surpass the baseline models that show the impact of large transformer-based models in this evaluation task. The models trained on heuristically generated implausible samples by random sentence/keyword substitutions show the lowest correlations. The main reason for such weakness is the huge dissimilarity of heuristically generated training data and machine-generated test data, which has a significant negative impact on the model's performance. The positive impact of UNION\_Data is visible in Table \ref{models_corr}. It demonstrates that the construction of implausible stories based on a more complete set of heuristic alterations yields better training data but still has its own shortcomings. This could be due to fact that text-level manipulations introduce artifacts that break the naturalness of the texts and have quite different styles compared to machine-generated implausible texts.

The superiority of RoBERTA and Longformer models trained on \textsc{ManPlts} and \textsc{AF\_ManPlts} datasets show the effectiveness of our proposed plot manipulation technique in enhancing the similarity between the training and test data. Adversarial filtering technique further helps to increase the quality of negative samples and generate better implausible machine-generated texts, which consequently improves the accuracy of evaluation. By applying hypothesis testing to compare the metrics correlations with human scores \cite{diedenhofen2015cocor}, we verify that these improvements are statistically significant (p<.05). We also note that the correlations between plot manipulation-based metrics and human evaluation are much higher in WP dataset. This could result from the limited ability of the current generative models to generate plausible long stories, thus making them easily distinguishable both by humans and automated metrics.

\begin{table}[t]
\begin{center}
\small
\begin{tabular}{lcccc}
\toprule
 \multirow{2}{*}{\textbf{Dataset}} &  \multicolumn{2}{c}{\bf ROC $\rightarrow$ WP} & \multicolumn{2}{c}{\bf WP $\rightarrow$ ROC } \\ \\ 
 & \bm{$\rho$} &  \bm{$\tau$}  & \bm{$\rho$} &  \bm{$\tau$} \\\midrule
  \textsc{UNION\_Data}
   &  0.17 &  0.15 & 0.12 & 0.07\\ 
\midrule
  \textsc{ManPlts}  
    & 0.57 & 0.39  & 0.23 & 0.16 \\
\midrule
  \textsc{AF\_ManPlts}
     & \bf 0.60 &  \bf  0.42 &  \bf 0.26 & \bf 0.18\\
  \bottomrule 
\end{tabular}
\end{center}
\caption{Correlation of plausibility metrics with human judgements. Arrow shows the train and test data used for examining metrics robustness.}
\label{data_drift}
% \vspace{0.1em}
\end{table}

One of the desirable features of automated evaluation metrics for story generation is their generalizability or robustness to different datasets \cite{sellam2020bleurt, Guan2020UNIONAU}. The dataset shifting robustness shows the metric's success in accurately evaluating texts in different datasets. We examine the robustness of metrics by leveraging ROC and WP as two distributionally different types of stories datasets. We train models on various training data constructed from negative sampling techniques in ROC dataset and test them on human scores collected through AMT experiments conducted on WP dataset (\textbf{ROC} \bm{$\rightarrow$} \textbf{WP}) and vice versa (\textbf{WP} \bm{$\rightarrow$} \textbf{ROC}). In Table \ref{data_drift}, we show the robustness of fine-tuned language models trained on the last three datasets of Table \ref{models_corr} as the best performing models in comparison to models trained on sentence and keyword substitutions.
According to Table \ref{data_drift}, the correlation drops due to the quite different structure of two datasets. RoBERTa/Longformer models fine-tuned on \textsc{AF\_ManPlts} in ROC/WP datasets and subsequently tested on WP/ROC dataset have the highest correlations with human judgments and can be generalized well on two datasets. The data shifting from ROC to WP better preserves the performance of metrics rather than the counterpart shifting. The reason for correlation decline of models trained on WP and tested on ROC could be the format of implausible texts in WP that could not be found in ROC data since the stories are shorter in this data and the reason for implausibility is fewer.
%This indicates that negative samples generated based on manipulated plots and adversarially filtered examples provide high-quality training data which helps the evaluator to generalize well on different datasets.

\begin{table}
\centering
\small
\begin{tabular}{lcc}
\hline
\textbf{Dataset} & \bm{$\rho$} & \bm{$\tau$}\\
\hline
\textsc{ManPlts-ReOrder} & 0.65 & 0.42\\
\textsc{ManPlts-ContInser} & 0.69  & 0.45\\
\textsc{ManPlts-RepeatInser} & 0.67 & 0.43\\ 
\textsc{ManPlts-RandSub} & 0.68 & 0.45\\ 
\hline
\end{tabular}
\caption{Correlations of Longformer model fine-tuned on plot-level manipulated datasets with one specific excluded technique.}
\label{tab:ablation_study}
\vspace{-1.5em}
\end{table}

\subsubsection{Ablation Study}
The positive impact of plot-level manipulations in precisely evaluating the plausibility can be assessed with regard to the four different manipulation techniques. We conduct an ablation study on WP dataset to examine each manipulation technique's impact separately. We construct different training data each time by excluding one of the manipulation techniques and generating a new set of negative samples. Then we fine-tune Longformer on all these training datasets with different negative samples and compute the correlation of the fine-tuned Longformer as the evaluation metric with human judgments. 
%We fine-tune Longformer on different datasets by excluding one specific manipulation technique and generating negative samples, then computing the correlation of the fine-tune Longformer model as the evaluation metric with human judgments. 

The lower correlations shown in Table \ref{tab:ablation_study} in comparison to Table \ref{models_corr} illustrate the harms that the elimination of each of the proposed approaches from the construction of training data could cause. This attests to the effectiveness of all proposed manipulation techniques in the generation of higher quality training data and subsequently resulting in more accurate evaluation metrics.
%decreases the correlations four techniques used for manipulation are quite effective in reaching accurate evaluation metric. 

As this table demonstrates, the correlation drops the most by ablating the reordering and repeating plots, which shows that they are the major problems in generating long texts by language models and have the most significant role in constructing high-quality implausible samples and consequently accurate evaluation metrics.

\section{Conclusion} \label{label:conclusion}
Automatic plausibility evaluation models that are trained on heuristically generated data show low correlation with human judgement. We address this issue by creating a better quality set of implausible texts. In contrast to existing methods that modify text at token level, our approach introduces incoherence sources at a more abstract plot level, which helps to guide the generative model conditioned on those manipulated plots to generate negative samples that are more similar  to machine-generated incoherent texts. We further improve the data quality by applying adversarial filtering to select more challenging and refined negative samples. Our experiments demonstrate that negative examples generated according to the proposed method result in more realistic implausible texts and consequently lead to more accurate evaluation metrics that have higher correlation with human judgement. 

\section{Ethics}
All co-authors of this work totally understand and agree with \textit{ACM Code of Ethics} and its importance in expressing the conscience of the profession. We ensure this work is compatible with the provided code, specifically in the terms of providing non-offensive dataset construction. 
%Two main steps in providing coherence evaluation metrics are: 1) training data construction and 2) testing data collection. 

\textbf{1) training data construction} 
%A key determinant in the success of evaluation metrics is the high-quality training data. 
In our approach, we use BART model conditioned on manipulated story plots to construct implausible samples that better reflect the implausibility in generation models. The main concern that arises here is the probability of generating abusive language samples from manipulated plots. Indeed, these plots origin from human-written stories without abusive languages provided by \cite{mostafazadeh2016corpus, fan2018hierarchical} where users are not allowed to write profanity and inappropriate content. Accordingly, our manipulated version of plots and the BART model conditioned on them generate samples unlikely to contain strong biases or abusive content. It is noteworthy to mention that even the source plots are relatively benign, the process of altering them would have the possibility of creating objectionable texts. Other potential attack could be the dual-usage of metrics by augmenting offensive language texts as plausible samples. This would harm the underlying tasks such as story generation models to be encouraged to generate inappropriate stories. Such attacks can be identified and dissolved by security trended studies which are out of this work's scope.

\textbf{1) testing data collection} We collect human judgments by conducting Amazon Mechanical Turk (AMT) experiments that are leveraged to compare the accuracy of trained metrics in terms of their correlations with human scores. The conducted AMT does not disrupt user privacy as we do not contain their personal information. This fades the possibility of any gender bias problems and IRB approval needs. Annotators were asked to rate the coherence of stories in each HIT page of AMT in the range of 0 up to 5. We fairly compensated annotators. The average time of annotating each HIT in AMT was 25 minutes (including three stories for evaluation and their explanations), and according to the per hour wage of \$13, we fairly paid them \$6 per HIT.

This work targets the NLP open-domain generation community. Our metrics establish the main basis to achieve higher-quality generations by automatically assess the outputs and save time, cost, and human efforts. We don't anticipate specific failure modes in our work since the provided approach's success has been investigated through a comprehensive set of comparisons with other existing metrics.
\section*{Acknowledgment} \label{label:acknowledgment}
This work is supported by the CwC program under the Contract W911NF-15-1-0543 with the US Defense Advanced Research Projects Agency (DARPA). We would like to thank the anonymous reviewers for their helpful comments and the members of PLUSlab from USC/UCLA, Shushan Arakelyan, and Ninareh Mehrabi for their constructive feedback.

\bibliography{anthology,custom}
\bibliographystyle{acl_natbib}
% \input{appendix}
% \appendix

% \section{Example Appendix}
% \label{sec:appendix}

% This is an appendix.

\end{document}